# Default Reasoning and the Transferable Belief Model.

Philippe Smets and Yen Teh Hsia[1]
IRIDIA, Université Libre de Bruxelles
50, av. F. Roosevelt. CP194/6, 1050
Bruxelles. Belgium

## 1. Introduction.

Inappropriate use of Dempster's rule of combination has led some authors to reject the Dempster-Shafer model, arguing that it leads to supposedly unacceptable conclusions when default is involved. The most classic example is that of the penguin Tweety. This paper will successively present:

1.  the origin of the miss-management of the Tweety example;
2.  the two types of default;
3.  the correct solution for both types based on the transferable belief model (our interpretation of the Dempster-Shafer model (Shafer 1976, Smets 1988));

Except when explicitly stated, all belief functions used in this paper are simple support functions, i.e. belief functions for which only one proposition (the focus) of the frame of discernment receives a positive basic belief mass with the remaining mass being given to the tautology. Each belief function will be described by its focus and the weight of the focus (e.g. $m(A)=.9$). Computation of the basic belief masses are always performed by vacuously extending each belief function to the product space built from all variables involved, combining them on that space by Dempster's rule of combination, and projecting the result to the space corresponding to each individual variable.

## 2. The Miss-Managment of Tweety

Users of Dempster-Shafer theory have sometimes treated the Tweety example as follows:
- they create three propositional primitives B, P and F (for Bird, Penguin and Fly).
- they create the implication $B \supset F$ ("if bird then fly").

Of course the aim of this implication formula is to translate the assertion "generally birds fly", "typically birds fly", "usually birds fly", "most birds fly"...into the belief function framework. To cope with these fine shades of meaning, one gives a less than one basic belief mass (e.g. .9) to the material implication $B \supset F$, i.e. $m(B \supset F) = .9$. When one learns that Tweety is a bird, another belief function $m(B)=1$ is added and the system deduces the combined belief $m(F)=.9$.

Then one learns that Tweety is in fact a penguin. Penguins are birds, so one introduces the belief $m(P \supset B)=1$. As usually penguins do not fly, one also introduces the belief $m(P \supset \neg F)=.95$. After introducing a fourth belief function $m(P)=1$, two belief functions are induced on F, one derived from the birdness of Tweety ($m(F)=.9$) and the other derived from the penguinness of Tweety ($m(\neg F)=.95$). By Dempster's rule of combination, one obtains the overall belief

[1] Research has partly been supported by the DRUMS (Defeasable reasoning and Uncertainty Management Systems) project funded by EEC grants under the ESPRIT II Basic Research Project 3085 and the Belgian national incentive-program for fundamental research in artificial intelligence

function with $m(F) = .31$, $m(\neg F) = .66$ and $m(F \vee \neg F) = .03$.

Of course the expected solution is $m(\neg F) = .95$ and $m(F \vee \neg F) = .05$ as Tweety is a penguin and the belief that Tweety flies given it is a penguin is .95. The origin of the error results from the non-applicability of Dempster's rule of combination. Indeed this rule can only be applied when the two pieces of evidence that induce the two belief functions $m_1(B \supset F)$ and $m_2(P \supset \neg F)$ are distinct.

In the present case, distinctness is not satisfied, because birds and penguins are related: birds are either penguin-birds (P) or non-penguin-birds (NPB): $B = P \vee NPB$. Hence $bel(F|B) = bel(F|P \vee NPB)$, so $bel(F|B)$ results from the disjunctive combination of two belief functions - $bel(F|P)$ and $bel(F|NPB)$ - induced by the two distinct pieces of evidence (P and NPB). Remember that Dempster's rule of combination concerns the conjunctive combination of two belief functions induced by the two distinct pieces of evidence as in:
$$bel(F|X \& Y) = bel(F|X) \oplus bel(F|Y)$$

In the present case, we are facing the problem of disjunctive combination of distinct pieces of evidence. We have shown in Smets (1978) that if P and NPB are two distinct pieces of evidence concerning our belief about F then
$bel(F|B) = bel(F|P \vee NPB)$
$\qquad = bel(F|P).bel(F|NPB)$

Even when this rule does not apply, $bel(F|B)$ is somehow related to $bel(F|P)$. Hence Dempster's rule of combination is not applicable in the above example, as the two pieces of evidence B and P are not distinct. We develop a more correct analysis below.

## 3. Two types of default.

A default may be written as A:B/C (Reiter 1980) where A, B and C are formulae. It is read "if A is true and B is consistent, then C is true". A is the prerequisite, B is the justification and C is the consequence of the default (Besnard 1989). A normal default (A:B/B) is a default in which the justification and the consequence are the same. It could also be written as $A:A \supset B/A \supset B$. Both forms of default yield the same results, while normal defaults without prerequisite $:A \supset B/A \supset B$ give different results. A:B/B is called a standard normal default (Touretzky 1984), whereas $:A \supset B/A \supset B$ is called a normal default without prerequisite (Poole 1985). The question as to which of the two types of default is appropriate depends, in fact, on the context. The difference between the two types of default can be seen in the case of modus tollens and case reasoning (Besnard 1989).
- With $:A \supset B/A \supset B$, modus tollens is applicable. From $\neg B$, one deduces $\neg A$ if the default can be applied. It is not the case with A:B/B.
- With $:A \supset X/A \supset X$ and $:B \supset X/B \supset X$ one infers X from $A \vee B$. It is not the case with A:X/X and B:X/X.

Some authors feel that modus tollens should be avoided as in the following example: if a person is kind, then he is popular and if a person is fat, then he is not popular. If John is fat, he is indeed not popular. But one would like to avoid the inference that John is not kind. Hence standard default should be used in this case. Sometimes default without prerequisite seems to be more appropriate, as in the following case: birds fly, bats fly, coco is either a bird or a bat, hence coco flies. The question as to which default is more appropriate is not discussed here. We only show how both types of default may be represented using belief functions.

## 4. Normal Defaults without Prerequisites ( $:B \supset F/B \supset F$ ).

Normal Defaults without Prerequisites is solved by the introduction of an auxiliary variable (as is done in McCarthy's anomaly theory). Let B, P and F represent Bird, Penguin and Fly respectively. We introduce the auxiliary binary variable TB which represents the concept of Typical-Bird. The default $:B \supset F/B \supset F$ is represented by $B\&TB \supset F$. The network in Figure 1 is the screen dump extracted from MacEvidence (Hsia and Shenoy 1989). In the inner rectangular area, there are three nodes B, TB and F and a link among them. This links represents our belief on the joint BxTBxF space. It is characterized here by the simple support function $m(B\&TB \supset F) = 1$. Furthermore, the fact that things are usually typical (in this case birds are usually typical) is encoded by the belief $m(TB) = .9$ (fed into MacEvidence through the pTB box).

Suppose we learn that Tweety is a bird. So we put $m(B) = 1$, and the system deduces $m(F) = .9$. Then we learn about penguin P. We must create new nodes and links that represent the new pieces of information we have about penguin (see the rectangular area in the centre of Figure 1). A penguin is a bird: $m(P \supset B)=1$. It is an atypical bird: $m(P \supset \neg TB)=1$. Let us accept that "usually penguins do not fly" (for the sake of generality, we assume the existence of SuperPenguin which flies), so $m(P\&TP \supset \neg F)=1$ and $m(TP)=.95$, where TP means Typical-Penguin and the "usuallity" level is quantified by .95. Since Tweety is a penguin, we put $m(P)=1$. The system deduces $m(B)=1$ and $m(\neg F)=.95$ as required. The inference from B to F is blocked, because TB is no longer true $(m(\neg TB)=1)$

Later, we learn that Tweety is SuperPenguin, and we add a new layer (outer rectangular area in Figure 1) with variable S (SuperPenguin), TS (Typical-SuperPenguin) and links $m(S \supset P)=1$, $m(S \supset \neg TP) = 1$, $m(S\&TS \supset F) = 1$ and $m(TS)= .99$ (since a SuperPenguin may have a broken wing). Table 1 summarizes some of the results one could derive according to which variables are instantiated (by giving a belief 1 in the variables).

As can be seen from Table 1, many back propagations occur. This is due to the fact that the prior beliefs about typicalities are fixed and cannot be modified. In the first line of Table 1, the $m(\neg P)=.9$ corresponds to the fact that usually birds are typical, hence by the link $P \supset \neg TB$, one gets the back propagation of .9 from $\neg TB$ to $\neg P$ (a result that might justify the rejection of this formulation). A nice result is that the truth of B (second line) does not modify our belief in P. The modus tollens is observed (the fifth line where $\neg F$ is true). $\neg F$ leads to $\neg B$. The modus tollens also explains the

¬S (i.e. it it is not SuperPenguin). Our belief in P remains unchanged.

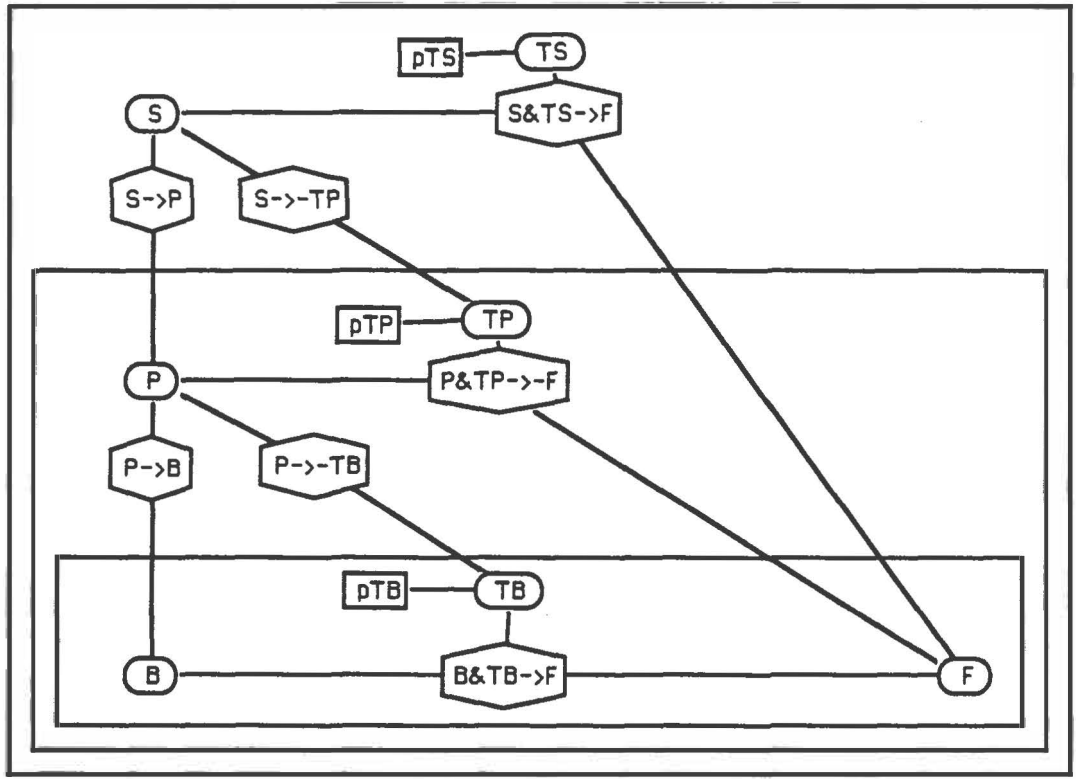

**Figure 1:** Analysis of the Tweety example with normal defaults without prerequisites. Oval nodes represent variables, six-sided polygons represent links (characterized by logical constraints).

|      | B      | P        | S        | F       |
|------|--------|----------|----------|---------|
| ?    | ?      | ¬P  .9   | ¬S  .995 | ?       |
| B    | B  1.  | ¬P  .9   | ¬S  .995 | F  .9   |
| P    | B  1.  | P   1.   | ¬S  .95  | ¬F .95  |
| S    | B  1.  | P   1.   | S   1.   | F  .99  |
| ¬F   | ¬B .9  | ¬P  .9   | ¬S  1.   | ¬F 1.   |
| B  F | B  1.  | ¬P  .995 | ¬S  .995 | F  1.   |
| B ¬F | B  1.  | ?        | ¬S  .999 | ¬F 1.   |
| P  F | B  1.  | P   1.   | ?        | F  1.   |
| B ¬P | B  1.  | ¬P  1.   | ¬S  1.   | F  .9   |

**Table 1:** Normal defaults without prerequisites. Content of this table describes the simple support functions induced on propositions B, P, S and F according to the instantiated variables (left column). Each pair corresponds to the focus and its weight. '?' represents the vacuous belief function.

The three rectangular areas in Figure 1 demonstrate the modular nature of our constructions. That is, whenever a new concept is introduced, previous variables and beliefs are kept unmodified; only new variables and new beliefs are added.

## 5. Standard Normal Defaults (B:F/F)

We now introduce the binary variables InB ("I learn that it is a bird"), TB (Typical-Bird), B (Bird) and F (Fly). When InB is instantiated (m(InB)=1), we deduce that it is indeed a bird (m(InB⊃B)=1) and it is "probably" a typical bird (m(InB⊃TB)=.9). We know that typical bird flies (m(B&TB⊃F)=1). The inner box of Figure 2 presents this model. After instantiating InB (m(InB)=1), we get m(B)=1, m(TB)=.9 and m(F)=.9. To know whether it is a bird, we check variable B (not InB, as the latter represents the learning of the fact that it is a bird). To show the difference between B and InB, suppose you are worried about the fact that birds lay eggs. So we create the variable "Eggs" and the link m(B⊃Eggs)=1. (and of course not m(InB⊃Eggs)=1: we don't get eggs because we learn that it is a bird; rather we get eggs because it is a bird).

Modus tollens is blocked in this formalism as m(¬F)=1 induces m(B∨¬B)=1, m(TB∨¬TB)=1 and m(¬InB)=.9. So one does not get any belief in ¬B from ¬F.

The general interpretation of the diagram presented in Figure 2 is: if you tell me that Tweety is a bird, I will conclude that it is a bird and probably a typical (.9) bird. I know that typical birds fly. So I believe at level .9 that Tweety flies.

Next, we introduce the concept of penguin (middle box in Figure 2). So we introduce the variables InP ("I learn that it is a penguin"), TP (Typical-Penguin as far as flying is concerned), XP (an auxiliary variable used to stop back propagation of TB to ¬P), and P (Penguin). The constraints are m(InP⊃TP) = .95, m(InP⊃P) = 1, m(InP⊃XP) = 1, m(P⊃B) = 1, m(P&XP⊃ ¬TB) = 1 and m(P&TP⊃¬F) = 1. Instantiating InP by m(InP)=1, we get m(¬F)=.95, m(B)=1 (even if m(InB)=1 had been instantiated).

When SuperPenguin is introduced, we create nodes and links in a way that is similar to what we did for Penguin. However, we also create S&XS⊃XP. It is needed in order to block a belief in TB when InB is instantiated. InS instantiates S, hence P by the link S⊃S. But the variable XP must also be instantiated to block the relation B to F at TB through the link P&XP⊃¬TB.

Table 2 summarizes some of the results one could derive according to which variables are instantiated after the introduction of SuperPenguin (which may have a broken wing). Note the difference between the ninth and tenth lines of Table 2. In the case of InB and ¬P, (i.e. we learn that Tweety is a bird and we know that Tweety is not a penguin), no back propagation occurs. Also note that both InB and InB&¬P lead to m(F) = .9. That is, ¬P does not modify our belief about F once B is learned. This would not have been the case if we had specified the rule that the only non typical birds are the penguins (by adding $m(\neg TB \supset P)=1$)). In fact, if we had added $m(\neg TB \supset P)=1$, InB&¬P would have led to $m(F)=1$ (i.e. we know that it is a bird and that it is not a penguin; hence it is a typical bird, and therefore it certainly flies, as the only exception is now impossible).

|         | B    | P     | S    | F      |
|---------|------|-------|------|--------|
| ?       | ?    | ?     | ?    | ?      |
| InB     | B 1. | ?     | ?    | F .9   |
| InP     | B 1. | P 1.  | ?    | ¬F .95 |
| InS     | B 1. | P 1.  | S 1. | F .99  |
| ¬F      | ?    | ?     | ?    | ¬F 1.  |
| InB F   | B 1. | ?     | ?    | F 1.   |
| InB ¬F  | B 1. | ?     | ?    | ¬F 1.  |
| InP F   | B 1. | P 1.  | ?    | F 1.   |
| InB ¬P  | B 1. | ¬P 1. | ¬S 1.| F .9   |
| InB InP | B 1. | P 1.  | ?    | ¬F .95 |
| InB InS | B 1. | P 1.  | S 1. | F .99  |

**Table 2:** Standard normal defaults. Content of this table describes the simple support functions induced on propositions B, P, S and F according to the instantiated variables (left hand column). Each pair corresponds to the focus and its weight. '?' represents the vacuous belief function.

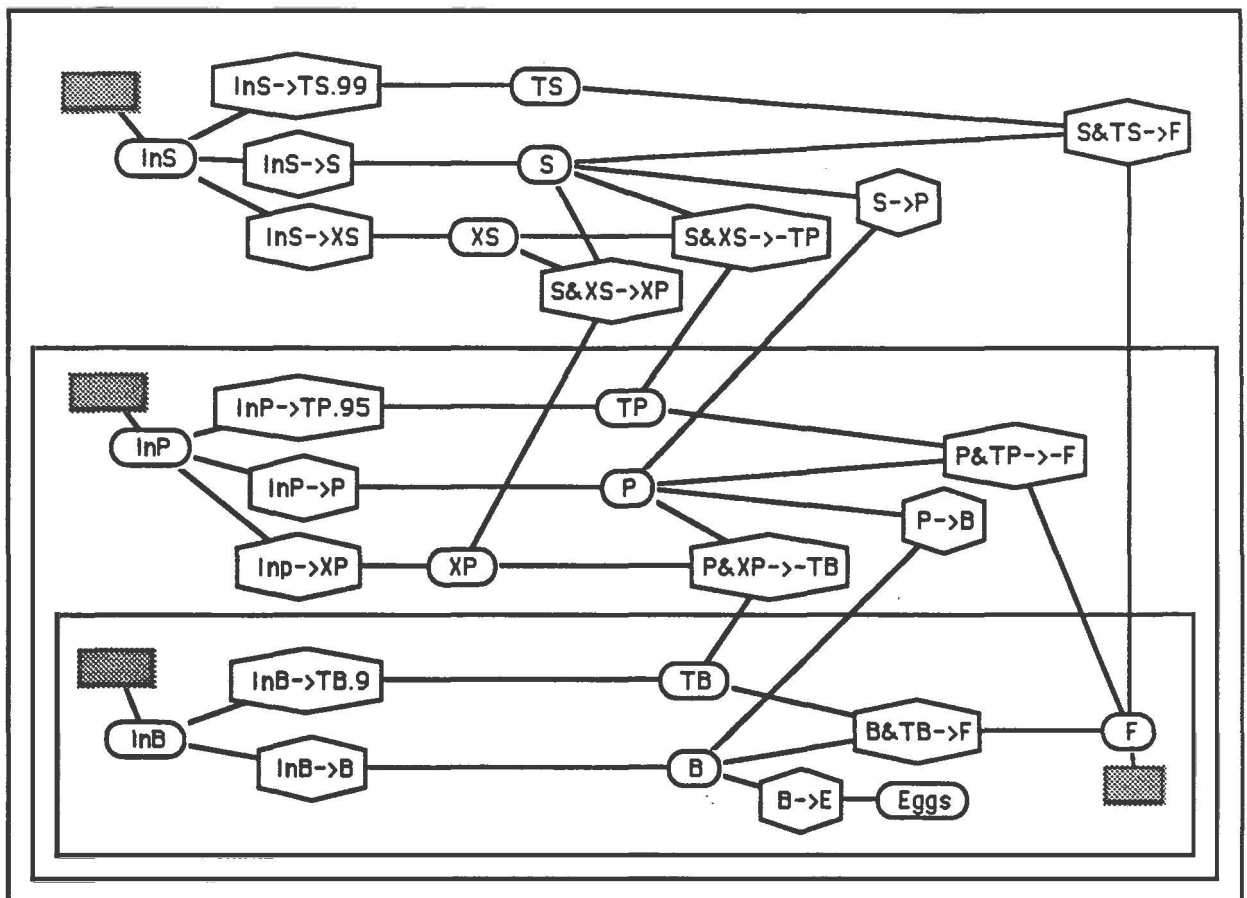

**Figure 2:** Analysis of the Tweety example with standard normal defaults. Oval nodes represent variables. Six-sided polygons represent links: those that include a number are weighted links (i.e. the implication receives the weight indicated), the others are logical links. Grey boxes are instantiating nodes (i.e. nodes through which the beliefs given to the variables can be initiated).

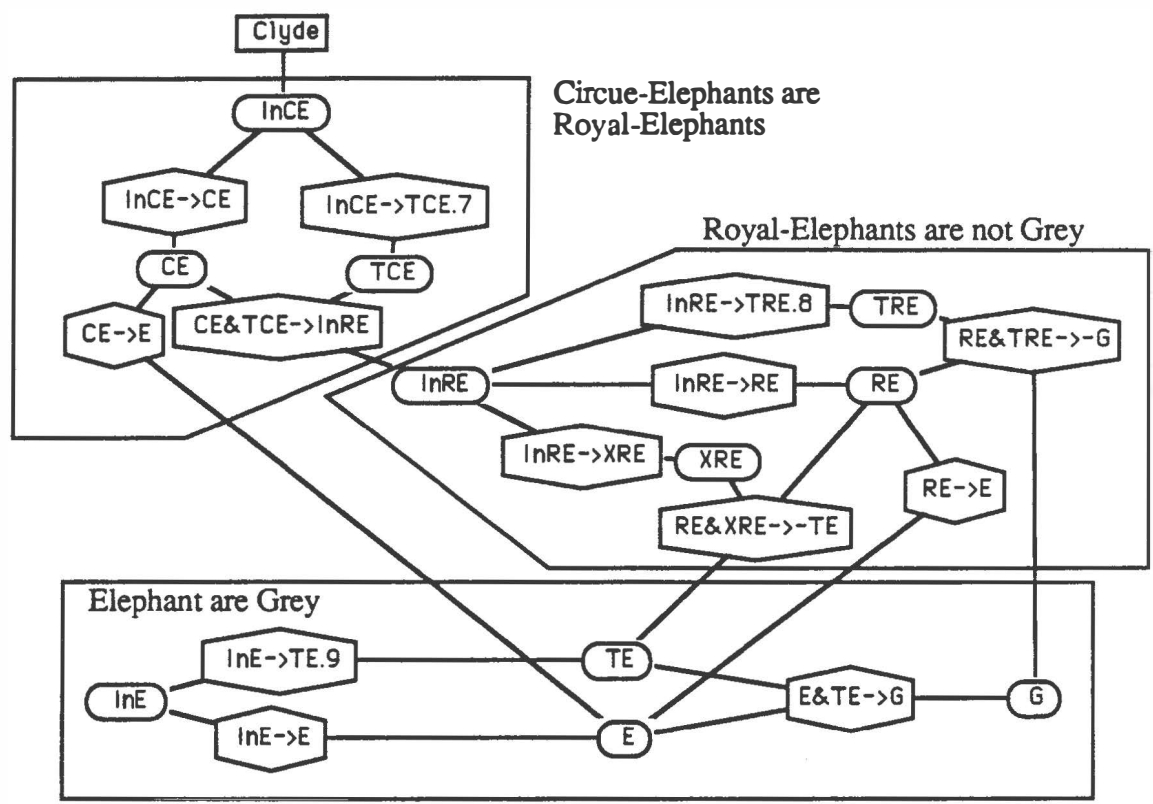

**Figure 3:** Clyde example.

The network in Figure 2 is cumbersome, but an auxiliary variable must be created each time we want to introduce a 'unidirectional implication', i.e. a rule that will propagate beliefs in one direction only. Should bidirectional propagations be acceptable, many auxiliary variables and links can be removed (the resulting network would then resemble Figure 1). That is, if we accept that the instantiation $m(InB)=1$ leads to $m(\neg P)=1$, etc, then the network becomes much simpler. We developed our formulation in order to show that it is possible to block all undesirable modus tollens. For instance, InB induces a belief in TB but leaves a vacuous belief on P, even though "usually $P \supset \neg TB$". Furthermore when both InB and InP (or InS) are instantiated, the flieness of Tweety is related only to its P (or S) status. Therefore our formulation satisfies the "preference by specificity" requirement (Moinard 1990).

## 6. Clyde Greyness.

As further illustration of the capacity of our construction, we now analyse the Clyde example. One has the standard normal defaults: Circus-Elephants are Royal-Elephants, Royal-Elephants are Elephants, Elephants are Grey, Royal-Elephants are not Grey, Clyde is an Circus-Elephant. What colour is he (with or without the link Circus-Elephants are Elephants)? Figure 3 presents the implementation of this problem in MacEvidence. It concludes that Clyde is not grey independently of the rule Circus-Elephants are Elephants. If one instantiates both InE and Clyde, the result is still Not Grey.

## 7. Conclusions.

We have developed two formulations that correspond respectively to two forms of normal defaults: $:A \supset B/A \supset B$ and $A:B/B$. Both formulations can be implemented in MacEvidence, a programme that runs on Macintosh and computes belief functions for any network. The differences between the two formulations can be found in their behaviour in back propagations. Modus tollens applies to the first formulation, but not to the second. The choice between the two formulations must be based on the requirements that must be satisfied in a specific application.

It is worth mentioning that our formulations might break down if used blindly when trying to derive the solutions in binary logic, i.e. by using only 0 and 1 weights. We shall now show that in fact the solutions found in binary logic are the limits of our solutions when the weights given to the 'typicality' variables tend to 1.

If we specify $m(TB)=1$ in section 4, then by instantiating $m(P)=1$, we get two strictly contradictory beliefs at the node TB: one comes from the prior $m(TB)=1$ and the other from P which induces $m(\neg TB)=1$. Nevertheless, if we initiate the pTB and pTP such that $m(TB)=x$ and $m(TP)=y$, then for all $x<1$ and $y<1$, $m(B)=1$ induces $m(F)=x$ and $m(P)=1$ induces $m(\neg F)=y$. So the limits for x and y approaching 1 are such that $m(B)=1$ induces $m(F)=1$ and $m(P)=1$ induces $m(\neg F)=1$.

Suppose that in our formulation for the standard normal defaults (see section 5), we give all the 'typicality' the weight 1. Deductions based on instantiations of InB, InP or InS are unaffected, except inasmuch as the deduced beliefs on F will have weights equal to 1. Problems appear, for instance, when both InB and $\neg F$ are instantiated, as the normalization phase of Dempster's rule of combination will not be possible at the F node. Nevertheless, for all values $x = m(InB \supset TB) < 1$, one gets $m(F) = x$. Hence the limit of $m(F)$ for x approaching 1 is also 1.

Our two formulations handle both hard and weak exceptions. Suppose x is A and usually A implies B and x is an exception. With hard exceptions, the implication link $A \supset B$ is blocked and we have $\neg B$. With weak exceptions, only the implication link $A \supset B$ is blocked. In Figures 1 and 2, we considered hard exceptions. To get weak exceptions, we simply remove the links $S\&TS \supset F$, $P\&TP \supset \neg F$.

Finally, the modularity of our formalisms is clear. Generalization is easy for other cases, e.g. the introduction of concepts such as 'strong wings' (in which case birds are more typical and the belief that they fly will be increased) or 'colibri' (in which case the non-typicality does not apply). The full power of the transferable belief model can be seen when the whole structure is generalized through the use of belief functions more elaborate than those used here. Imagine that when I learn InB, I am not sure that B is true as the source that told me InB might be unreliable. Many other generalizations can be described in a similar manner.

## Acknowledgments:
The authors are indebted to Ph. Besnard, R. Kennes, Y. Moinard, J.Pearl and A. Saffiotti for their useful and most stimulating comments.